\documentclass[letterpaper]{article}
\usepackage[preprint]{aaai2027}

\usepackage[hyphens]{url}
\usepackage{graphicx}
\usepackage{natbib}
\usepackage{caption}
\usepackage{amsmath}
\usepackage{amssymb}
\usepackage{algorithm}
\usepackage{algorithmic}
\usepackage{booktabs}
\usepackage{colortbl}

\definecolor{bestcell}{RGB}{169,203,216}
\definecolor{secondcell}{RGB}{210,224,202}
\newcommand{\best}[1]{\cellcolor{bestcell}#1}
\newcommand{\second}[1]{\cellcolor{secondcell}#1}
\newcommand{\legendbox}[2]{{\setlength{\fboxsep}{1.2pt}\colorbox{#1}{#2}}}

\newcommand{\model}{TacPAC}

\title{\model{}: Tactile Prediction and Real-Time Action Correction in World-Action Models for Contact-Rich Manipulation}
\author{
    Zipei Ma\textsuperscript{\rm 1,\rm 2,\rm 3},
    Xiaofei Wei\textsuperscript{\rm 1,\rm 2},
    Junzhe Jiang\textsuperscript{\rm 1},
    Shunlin Lu\textsuperscript{\rm 3},
    Li Zhang\textsuperscript{\rm 1,\rm 2}
}
\affiliations{
    \textsuperscript{\rm 1}School of Data Science, Fudan University
    \textsuperscript{\rm 2}Shanghai Innovation Institute
    \textsuperscript{\rm 3}NeoteAI
}

\begin{document}

\maketitle

\begin{abstract}
World-action models guide action generation with predicted future observations, but vision-centric
predictions miss the local contact cues that decide contact-rich manipulation.  However, naively
predicting future tactile observations as additional views recovers only a third of the achievable
gain in our experiments.  This gap reflects a timing mismatch: predictions precede execution,
while tactile feedback arrives during it.  We introduce TacPAC, which turns tactile prediction
into real-time action correction.  Once the base model has
planned an action chunk, TacPAC caches the predicted contact that plan was conditioned on together
with the plan's own representation, and a tactile expert reads
each newly observed tactile image against that cache to correct the actions not yet executed.
Feedback is thus interpreted against what the plan anticipated rather than in isolation, and one
correction is a single pass over that cache, $20.7\times$ cheaper than regenerating the chunk.  On
five real-robot tasks spanning precision insertion, fragile-object handling, object reorientation,
and long-horizon manipulation, TacPAC leads every task and raises the average from 22\% for its
vision-only base model to 64\%. Code is available at \url{https://github.com/LogosRoboticsGroup/TacPAC}.
\end{abstract}


\section{Introduction}
\label{sec:introduction}

\begin{figure*}[t]
\centering
\includegraphics[width=\textwidth]{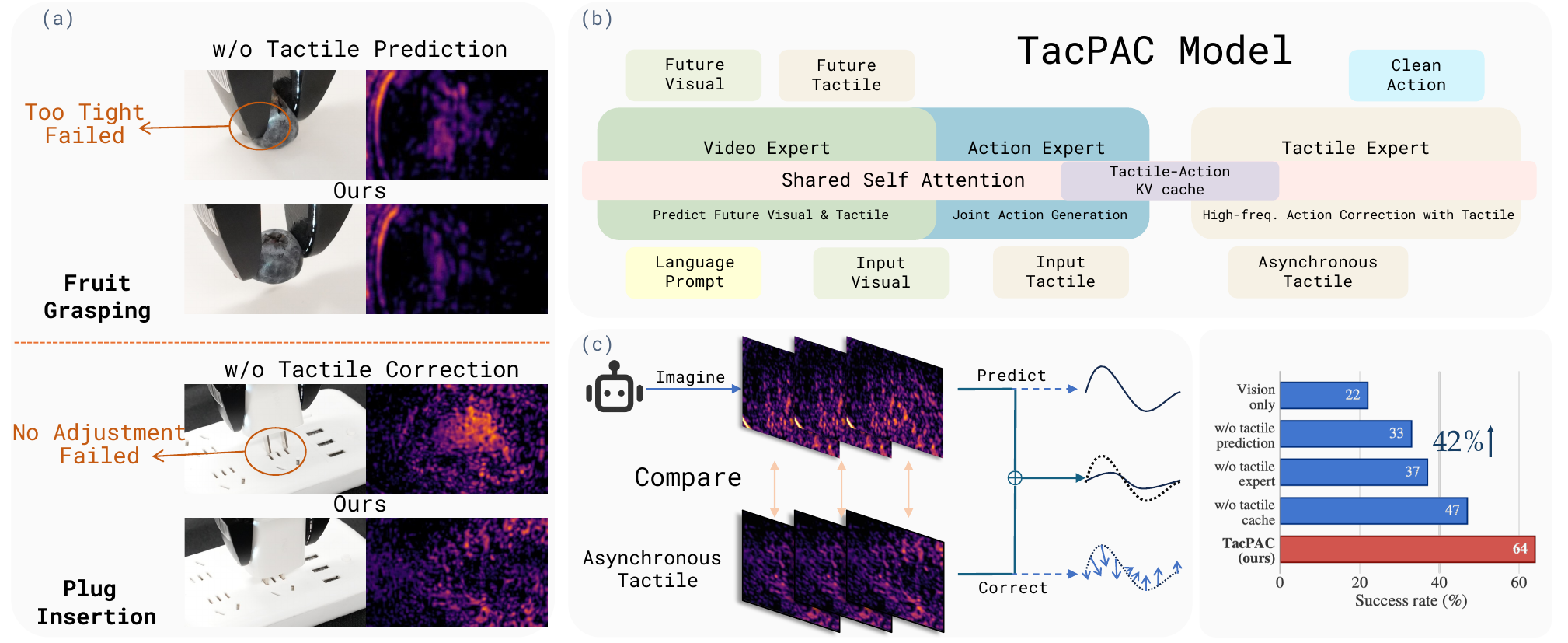}
\caption{\textbf{Tactile correction against the contact the plan expected.}  Tactile feedback
becomes actionable once the model knows what contact it expected.
\textbf{(a)}~Ablation experiment of real-world tasks, recorded from sensors during inference.  \textbf{(b)}~The base model generates future observations and the action chunk jointly,
after which the tactile expert corrects that chunk from the current tactile image and the cached tactile
and action states.  \textbf{(c)}~Our key insight: the correction compares
predicted against observed tactile, not the observation alone.
\textbf{(d)}~Success rate of each component ablation, averaged over the five real-robot
tasks.}
\label{fig:teaser}
\end{figure*}

Contact-rich manipulation requires a robot to adjust its actions from physical interaction, not
only from visible scene geometry.  Inserting a charger plug demands precise alignment under
constrained contact, and grasping a potato chip requires enough grip to lift it without breaking
it.  In both cases, images give essential global context but rarely resolve the local deformation,
incipient slip, or small insertion mismatch that decides the outcome.  Tactile sensing observes
these interaction changes directly and has shown value for manipulation
policies~\citep{george2025vital}.

Recent vision-language-action policies map multimodal context directly to action
chunks~\citep{brohan2023rt2,kim2024openvla,black2024pi0,intelligence2025pi_,zhao2023learning,yu2026wall,qwenvla,yuan2026qwen,wu2026foundation,zhang2026hy,team2026xiaomi},
whereas world-action models (WAMs) additionally predict future observations and let action tokens
attend to them, providing a predictive prior over the consequences of an
action~\citep{yuan2026fast,li2026causal,li2026wall,ye2026world,kim2026cosmos,ma2026dit4dit,zhang2026imagewam}.
This prior, however, is typically dominated by visual observations.  The resulting actions can
remain unreliable when task success depends on contact changes that are weakly expressed or
occluded in the image stream.

A natural remedy is to make the WAM predict tactile observations as well, treating tactile images
as additional views generated together with future
video~\citep{lou2026dreamtac,zheng2026omnivta,zang2026tacforesight,yuan2026vtam}.  In our
experiments this extension does improve task success, but it captures only about a third of the
gain that tactile sensing turns out to afford.  One reason may be that informative tactile changes
are temporally sparse, so a future-view objective need not make action generation sensitive to the
contact transitions that matter.  More fundamentally, a prediction made before execution cannot
revise actions that are already planned but not yet executed, so the key challenge is to connect
predicted tactile interaction with the tactile evidence acquired during execution.

We introduce \emph{\model{}}, which makes tactile prediction actionable through real-time action
correction (Fig.~\ref{fig:teaser}).  In the base model, a video expert predicts future visual and
tactile observations and an action expert attends to those predictions through
mixture-of-transformers attention~\citep{liang2025mixtureoftransformers}, so the planned action
chunk is conditioned on the contact it is expected to produce.  Once the chunk is planned, \model{}
caches that finished prediction and the plan itself as layer-wise attention states, and lets a
tactile expert read each newly observed tactile image against the cache.  The corrector therefore
sees not only what is being sensed but what the plan assumed would be sensed, and outputs a
per-step delta action for the steps not yet executed.  Because the base model runs once per chunk
while each correction is a single pass over the fixed cache, corrections keep up with the tactile
stream instead of waiting for the next plan.

We evaluate \model{} on five real-robot tasks---charger-plug insertion, multi-object fruit
transfer, fragile potato-chip transfer, empty-bottle uprighting, and expansion-card
insertion---in which tight tolerances and regulated contact forces turn small errors into jamming,
slip, or damage.  The
experiments answer four questions that mirror our claims: overall contact-rich capability (Q1), why
tactile prediction alone is insufficient (Q2), which components enable tactile action correction
(Q3), and whether the tactile expert is fast enough to correct a chunk in real time (Q4).
\model{} attains the highest success rate on all five tasks, improves the five-task average by 16
points over the strongest baseline, and issues each correction in $30.4$~ms against a reused cache.

Our main contributions are:
\textbf{(i)} We identify a practical limitation of tactile-predictive WAMs: predicting future tactile observations raises task success but leaves most of the achievable gain unrealized, as the prediction is fixed once the action chunk is planned;
\textbf{(ii)} We introduce \model{}, which caches the predicted contact a planned chunk was conditioned on together with the plan itself, so that current tactile feedback is interpreted against what the plan anticipated rather than in isolation;
\textbf{(iii)} We develop a single-pass tactile corrector, trained on the base model's own plans at densely sampled execution offsets, and evaluate the resulting system across five contact-rich manipulation tasks.

\section{Related work}
\label{sec:related_work}

\paragraph{World-action models.}
World-action models (WAMs) predict how observations evolve under actions and condition action
generation on that prediction. Recent systems adapt a pretrained video model into a policy by
encoding actions as latent frames of its diffusion process~\citep{kim2026cosmos}, cascade a video
and an action diffusion transformer so that intermediate denoising features rather than decoded
frames condition the action~\citep{ma2026dit4dit}, organize video--action learning around semantic
events instead of fixed-length chunks~\citep{li2026wall}, or learn from heterogeneous robot data to
transfer across tasks and embodiments~\citep{ye2026world}. Others question how much future
generation is needed at test time, retaining video co-training while skipping future prediction at
inference~\citep{yuan2026fast}. Our base model is architecturally closest to LingBot-VA, which
couples vision and action tokens in a Mixture-of-Transformers and re-conditions on ground-truth
observations during closed-loop rollout~\citep{li2026causal}. In all of these models, however, both
the predicted future and the feedback that refines it remain almost exclusively visual, so they
cannot anticipate the slip, incipient contact, or slight misalignment that decide contact-rich
manipulation. \model{} keeps this predictive backbone but makes both the prediction and the
feedback tactile.

\paragraph{Tactile manipulation and tactile-predictive models.}
Tactile sensing measures the contact state that vision cannot resolve, yet manipulation has used it
almost exclusively as a present-time signal.  High-resolution vision-based
sensors~\citep{lambeta2020digit,donlon2018gelslim} and self-supervised tactile
encoders~\citep{higuera2024sparsh,zhao2024t3,yang2024unitouch} feed loops closed on the current
reading: tactile servoing and compliance
control~\citep{lepora2021tactileservo,hogan2020tactiledexterity,tang2023industreal,zhang2023admittance},
visuo-tactile policies~\citep{lee2019visiontouch,li2022seehearfeel,huang2024threedvitac,george2025vital},
and VLAs conditioned on the current tactile or force
signal~\citep{xue2025reactive,huang2025tactilevla,zhang2025vtla,yu2025forcevla,yuan2026ftp}, for which
$\mathcal{N}_0$-Foundation consolidates hardware, data, and benchmarks~\citep{n0foundation2026}.  A
recent line instead brings tactile into \emph{predictive}
models~\citep{he2024foar,heng2025vitacformer,niu2026touchdreaming,zang2026tacforesight,zheng2026omnivta,ye2025dreamtacvla,n0twam2026},
but prediction and correction remain separate abilities: Dream-Tac anticipates contact yet commits to
the chunk in a single generative pass~\citep{lou2026dreamtac}, while T-Rex refines an
already-generated chunk from the live tactile stream without predicting tactile at all, so its
correction has no expectation to read against~\citep{niu2026trex}.  \model{} instead turns the
prediction into a live reference: the predicted tactile keys and values the chunk was planned under
are retained, and every incoming tactile image is read against them inside attention, so a correction
responds to how contact \emph{departs} from what the plan assumed rather than to the reading alone.
The distinction is measurable: at identical base model, corrector, and tactile input, blocking the
corrector's access to the cached prediction costs 17 points of average success
(Sec.~\ref{sec:exp_ablation}).

\section{Method}
\label{sec:method}

\subsection{Problem formulation}
\label{sec:method_problem}

The policy is conditioned on a task instruction, the current visual observation, and the
proprioceptive state.  At every time step $t$ the tactile sensor additionally provides a raw image
$\mathbf{I}_t$ of its contact surface.  The policy generates an action chunk
$\mathbf{a}=\{\mathbf{a}_i\}_{i=0}^{H-1}$ of horizon $H$ together with future visual and tactile
observations.  Since both modalities arrive as images, we tile them into one video stream encoded
into a single latent $\mathbf{z}$: each tactile sensor enters as an additional view alongside the
cameras, so predicting future tactile observations is part of predicting future video.

Both observation and action generation use conditional flow matching.  For a clean sample
$\mathbf{x}$ and Gaussian noise $\boldsymbol{\epsilon}$, we construct

\begin{equation}
\mathbf{x}_{\tau}=(1-\tau)\mathbf{x}+\tau\boldsymbol{\epsilon},
\qquad \tau\in[0,1],
\end{equation}
with target velocity $\mathbf{u}=\boldsymbol{\epsilon}-\mathbf{x}$.  The observed first frame is
kept clean, whereas future observation latents and the action chunk start from noise.  The flow
time $\tau$ indexes this generation path alone and is unrelated to the physical time $t$ at which
observations arrive.

\model{} builds on a tactile-predictive world-action model: a video expert and an action expert
organized as a Mixture-of-Transformers (MoT)~\citep{liang2025mixtureoftransformers}, which predict
the velocities for the noised video latent $\mathbf{z}_{\tau}$ and action chunk $\mathbf{a}_{\tau}$
and interact through layer-wise attention over the concatenated token sequence.  An asymmetric mask
lets every action query attend to all video and action tokens, whereas video queries attend only to
video tokens, so predicted observations condition action generation while noisy actions never
disturb observation prediction.  Because that video stream includes the tactile views, tactile
tokens play two roles at once: they are part of what the model generates, and, since action queries
attend to them at every layer, they are also part of what the chunk is generated from.  The chunk
is thus conditioned on the contact the model predicts its own actions will produce.  We call this
model the \emph{base model}.  Its estimate of the clean chunk, written $\hat{\mathbf{a}}$, is the
\emph{planned action chunk}: the actions committed before any execution feedback.

\begin{figure*}[t]
\centering
\includegraphics[width=\textwidth]{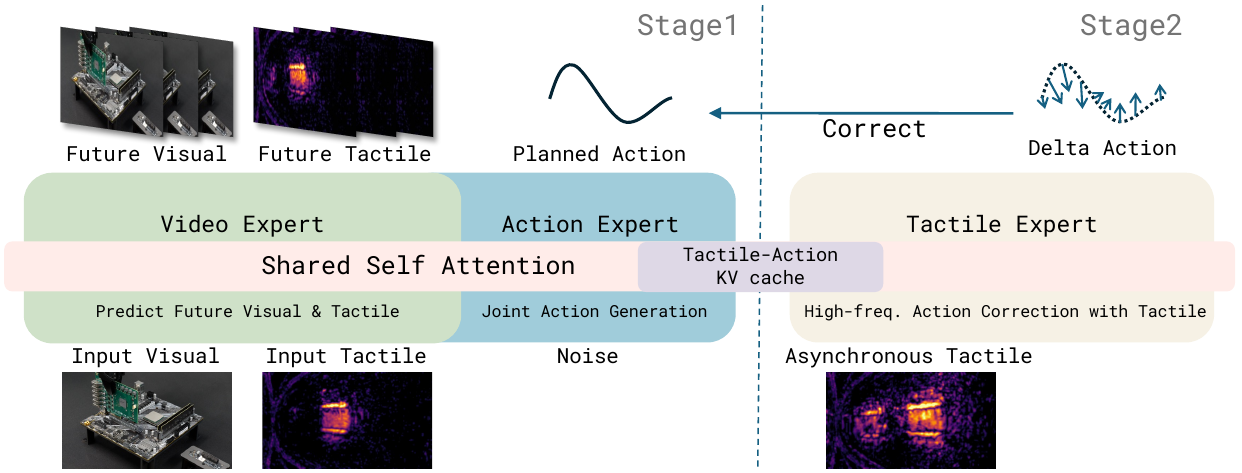}
\caption{\textbf{Overview of \model{}.}  Our method uses a three-expert Mixture-of-Transformers
architecture: a video expert predicts future visual and tactile observations, an action expert
conditions on these predictions to generate the planned action chunk $\hat{\mathbf{a}}$, and a
tactile expert outputs a delta action $\Delta\mathbf{a}$ that corrects $\hat{\mathbf{a}}$ from the
cached tactile--action KV and the current tactile observation.  Since the cache is reused, this
correction runs asynchronously at the tactile frame rate without rerunning the other two experts.
Stage~1 trains the video and action experts. Stage~2 freezes them and trains only the tactile
expert.}
\label{fig:pipeline}
\end{figure*}

That conditioning, however, is fixed once the chunk is planned: the anticipated contact is an
assumption made before execution, and nothing revises the plan when the contact that actually
occurs departs from it.  \model{} adds a correction path for exactly this situation
(Fig.~\ref{fig:pipeline}).  Sec.~\ref{sec:method_correction} describes the cache and the tactile
expert that reads it, while Sec.~\ref{sec:method_optimization} describes the two training stages and
inference.

\subsection{Prediction-grounded tactile correction}
\label{sec:method_correction}

\paragraph{Motivation.}
Correcting an action chunk from tactile feedback requires knowing not only what is being sensed
but what should have been sensed.  A corrector that sees only the current tactile image can tell
that the fingers are loaded, but not whether that load is the one the chunk was planned for.  The
same reading can call for opposite responses depending on whether contact was expected at this
execution stage.  In the base model this expectation is available: the chunk was denoised jointly
with a predicted future tactile stream and attends to it at every layer, so an error in the
prediction propagates into a correlated error in the chunk.  We therefore hand the corrector that
prediction together with the chunk's own internal representation, so that a new tactile image is
read against the contact the chunk anticipated.

\paragraph{Caching the prediction and the plan.}
Once the base model has completed the denoising schedule, we run one additional pass over the
denoised video latent and the planned chunk $\hat{\mathbf{a}}$ at flow time $\tau=0$ and store the
per-layer attention keys and values it produces.  We pay for this extra pass rather than reusing
the keys already produced by the final denoising step, ensuring that the resulting cache represents
a clean prediction rather than a still-noisy one.  The cache $\mathcal{C}$
retains two groups: the tactile views, which encode the contact the model predicted across the
horizon, and the action tokens, which encode the planned chunk as the model represented it while
attending to that prediction.  Visual keys are not retained: what a correction must resolve is a
contact that can change within a few action steps, whereas the scene it takes place in changes
little over a chunk and is already carried by both stored groups, which were computed while
attending to it.  The cache is built once per chunk and read unchanged by every subsequent
correction.

\paragraph{Tactile expert.}
The tactile expert is a single-pass corrector.  When a tactile image arrives after $m$ actions of
the chunk have been executed, its query sequence holds the current tactile image $\mathbf{I}_m$ and
an updated version of the planned chunk $\hat{\mathbf{a}}$.  The image is encoded by the same frozen
video tokenizer and tactile patch embedding that produced the cached tactile keys, allowing it to
enter their embedding space.  For the updated chunk, we replace the first $m$ entries with the
commands actually sent to the robot, retain the base model's planned values for the remaining
entries, and augment each entry with a learned status embedding indicating whether the corresponding
action has been executed or not.  Every layer attends these queries jointly to
$\mathcal{C}$ and to the expert's own
tokens. The expert outputs a per-step delta action

\begin{equation}
\Delta\mathbf{a}
=f\!\left(\mathbf{I}_m,\,\hat{\mathbf{a}},\,m,\,\mathcal{C}\right),
\end{equation}
which is added to $\hat{\mathbf{a}}$ to give the corrected chunk $\hat{\mathbf{a}}+\Delta\mathbf{a}$.
The expert shares the depth and attention geometry of the base experts so that it can consume their
per-layer cache.  The instruction reaches it only through those keys, which were produced under text
conditioning.  It is conditioned on the execution offset $m$ rather than on a flow time: the offset
places the arriving tactile frame at a stage of the plan, fixing which part of the cached
prediction it should be read against and how much of the chunk remains revisable.

\paragraph{Technical advantage.}
Cache-based correction offers two practical benefits. First, each correction requires only one forward pass of
the tactile expert over the tactile tokens and $H$ action tokens, attending to a fixed cache. It does not revisit
the video stream or denoising schedule. By contrast, refreshing predicted contact would rerun the video expert
over the full multi-view canvas, nearly matching the cost of a new plan. Cache reuse therefore supports corrections
at the tactile-frame arrival rate. Second, each correction is added to the original planned chunk rather than the
previously corrected one. The expert still reads the commands actually sent as its executed prefix, but its new
delta replaces the previous correction on the remaining steps. Thus, a spurious correction does not propagate to
later updates. This behavior is learned through attention without explicitly comparing predicted and observed
tactile images. Component ablations in Sec.~\ref{sec:experiments} evaluate the benefit of the cached prediction
over using only the current tactile image.

\subsection{Training and inference}
\label{sec:method_optimization}

\begin{algorithm}[t]
\caption{\textbf{Training and inference for tactile-expert.}}
\label{alg:tactile}
{\raggedright\textbf{Training} --- Tactile-expert update, base model frozen\par}
\begin{algorithmic}[1]
\REQUIRE ground-truth chunk $\mathbf{a}$ with tactile images $\{\mathbf{I}_t\}$, horizon $H$,
offsets per chunk $K$
\STATE Plan $\hat{\mathbf{a}}$ with the frozen base model and prefill $\mathcal{C}$,
\FOR{$k=1,\dots,K$}
\STATE Sample an offset $m\sim\mathrm{Unif}\{0,\dots,H{-}1\}$,
\STATE Overwrite the first $m$ entries of $\hat{\mathbf{a}}$ with ground truth,
\STATE Compute $\Delta\mathbf{a}\leftarrow f(\mathbf{I}_m,\hat{\mathbf{a}},m,\mathcal{C})$ and
accumulate \eqref{eq:tactile_loss} for $i\geq m$,
\ENDFOR
\end{algorithmic}
\vspace{4pt}
{\raggedright\textbf{Inference} --- One action chunk\par}
\begin{algorithmic}[1]
\STATE The base model completes denoising and returns $\hat{\mathbf{a}}$
\COMMENT{robot execution starts here},
\STATE Prefill $\mathcal{C}$ from the denoised video latent and $\hat{\mathbf{a}}$,
\WHILE{the chunk is not exhausted}
\STATE A tactile frame arrives at offset $m$: $\Delta\mathbf{a}\leftarrow
f(\mathbf{I}_m,\hat{\mathbf{a}},m,\mathcal{C})$,
\STATE Commit $\hat{\mathbf{a}}_i+\Delta\mathbf{a}_i$ for $i\geq m+d$,
\ENDWHILE
\end{algorithmic}
\end{algorithm}

\paragraph{Stage 1: base model.}
We first train the video and action experts over the complete flow interval, with the first
visual--tactile frame as the observation condition and independently noised future latents and
actions processed jointly by MoT.  The objective is a weighted sum of a video and an action
flow-matching loss: the video term covers both the future visual and the future tactile latents,
and the action term is applied in action space after converting the predicted velocity to a
clean-action estimate.  Stage~1 yields a base model that plans action chunks without the tactile
expert.

\paragraph{Stage 2: tactile expert.}
Stage~2 freezes everything trained in Stage~1 and optimizes only the tactile expert, initialized
from the trained action expert so that it starts from a module that already knows how to read the
base model's per-layer states.  Each update reproduces the deployment path
(Algorithm~\ref{alg:tactile}, \emph{Training}): the frozen base model plans a chunk under the
denoising schedule used at test time, and the prefill of Sec.~\ref{sec:method_correction} builds
$\mathcal{C}$ from that chunk.  Training on the model's own plan rather than on the ground-truth
chunk is what keeps the two paths consistent, since the quantity the expert must correct is
precisely the gap between what the base model plans and what the task requires.

The dataloader supplies several execution offsets per training episode, each paired with the
tactile image recorded at that offset.  For an offset $m$, the executed prefix is filled with the
ground-truth actions---the commands that produced that tactile image---while the remaining steps
keep their planned values, and the expert is supervised only where a correction can still be
applied:

\begin{equation}
\mathcal{L}_{\mathrm{tac}}
=\frac{1}{H-m}\sum_{i\geq m}
\left\lVert\hat{\mathbf{a}}_i+\Delta\mathbf{a}_i-\mathbf{a}_i\right\rVert^{2}.
\label{eq:tactile_loss}
\end{equation}

The delta action is thus supervised to be exactly the residual between the planned chunk and the
ground-truth actions on the unexecuted suffix, conditioned on the tactile evidence available at
that point of execution.  Sampling offsets densely across the horizon exposes the expert to every
execution stage, and because one frozen plan and one prefill are shared by all offsets drawn
for the same episode, this supervision costs little more than a single base forward pass.

\begin{figure}[t]
\centering
\includegraphics[width=\linewidth]{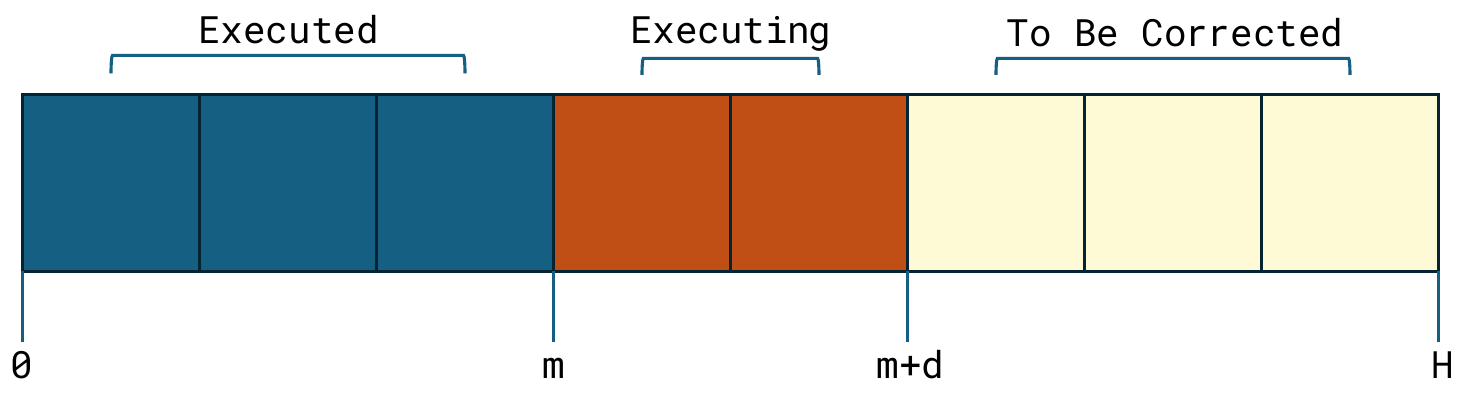}
\caption{\textbf{Asynchronous inference of the tactile expert.}  A tactile frame at chunk offset
$m$ triggers a delta action $\Delta\mathbf{a}$ that arrives $d$ steps later: steps before $m$
(blue) already ran, steps in $[m,\,m{+}d)$ (orange) run while $\Delta\mathbf{a}$ computes, and only
steps from $m{+}d$ to $H$ (yellow) apply it.}
\label{fig:infer}
\end{figure}

\paragraph{Inference.}
A chunk is served in three steps (Algorithm~\ref{alg:tactile}, \emph{Inference}).  The robot begins
executing the action chunk as soon as the base model generates it.  The prefill is then computed
while the first actions are already in flight.  Every tactile frame that arrives thereafter
triggers one correction against the active chunk, and the base model runs again only once the
horizon is exhausted.

Because the arm keeps moving while a correction computes, the correction cannot be applied at the
offset it was conditioned on: by the time it returns, those commands have been sent.  Each
correction is therefore conditioned at the offset $m$ at which its tactile frame was taken---the
pairing the expert was trained on---but written back only from $m+d$ onward, where $d$ is the
largest correction latency observed in a short recent window, expressed in action steps.  Steps
inside that window keep the values most recently committed for them, and already-executed actions
are immutable (Fig.~\ref{fig:infer}).  In practice this window is nearly empty: a correction
returns within about one action step of the tactile frame that triggered it, so $d$ is zero for
most of execution and each correction rewrites almost the entire unexecuted remainder of the chunk.
Sec.~\ref{sec:experiments} reports the measured latency, the achieved correction rate, and their
effect on task success.

\section{Experiments}
\label{sec:experiments}

\begin{figure*}[t]
\centering
\includegraphics[width=\textwidth]{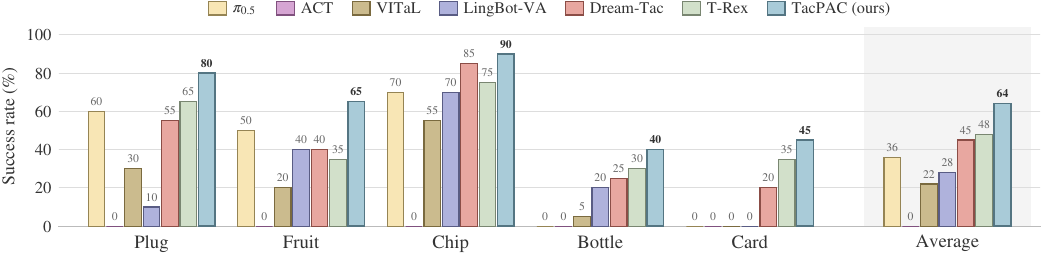}
\caption{\textbf{Main results on the five contact-rich real-robot tasks.}  Per-task and average
episode success rates, 20 trials per method per task.  \model{} attains the highest success rate on
every task.}
\label{fig:main_results_bar}
\end{figure*}

\begin{table*}[t]
\centering
\caption{\textbf{Component ablation of the tactile pathway.}  Each row cuts one pathway from
tactile sensing to action---tactile \emph{prediction}, execution-time \emph{correction}, or the
corrector's \emph{cache} access (--: no corrector).  Best \legendbox{bestcell}{blue}, second best
\legendbox{secondcell}{green}.}
\label{tab:component_ablation}
\small
\setlength{\tabcolsep}{4pt}
\begin{tabular}{lccccccccc}
\toprule
& \multicolumn{3}{c}{Tactile pathway} & \multicolumn{6}{c}{Success rate (\%) $\uparrow$} \\
\cmidrule(lr){2-4} \cmidrule(lr){5-10}
Variant & Prediction & Correction & Cache & Plug & Fruit & Chip & Bottle & Card & Average \\
\midrule
Vision only (w/o tactile) & $\times$ & $\times$ & -- & 15 & 30 & 60 & 5 & 0 & 22 \\
w/o tactile expert (prediction only) & \checkmark & $\times$ & -- & 35 & \second{50} & 65 & 20 & 15 & 37 \\
w/o tactile prediction & $\times$ & \checkmark & $\times$ & \second{40} & 25 & 45 & \second{30} & 25 & 33 \\
w/o tactile cache & \checkmark & \checkmark & $\times$ & \second{40} & \second{50} & \second{75} & \best{40} & \second{30} & \second{47} \\
\textbf{\model{} (Ours)} & \checkmark & \checkmark & \checkmark & \best{\textbf{80}} & \best{\textbf{65}} & \best{\textbf{90}} & \best{\textbf{40}} & \best{\textbf{45}} & \best{\textbf{64}} \\
\bottomrule
\end{tabular}
\end{table*}

We structure the evaluation around four questions: (\textbf{Q1}) overall performance on contact-rich
manipulation, (\textbf{Q2}) why tactile prediction alone leaves a large gap to the full system,
(\textbf{Q3}) which of tactile input, tactile prediction, the tactile expert, and the cached
prediction that expert reads are necessary, and (\textbf{Q4}) whether the tactile expert is fast
enough to correct a chunk in real time.  Q1 is reported as results and analysis
(Sec.~\ref{sec:exp_results}), while Q2--Q4 form the ablation study (Sec.~\ref{sec:exp_ablation})
that isolates individual design choices.

\subsection{Experimental setup}
\label{sec:exp_setup}

\paragraph{Robot and observation--action interface.}
We evaluate on a single-arm Flexiv Rizon~4 whose gripper integrates two InTac~S1 tactile sensors,
with a wrist-mounted Intel RealSense D405 and a third-person Intel RealSense D435i.  \model{} takes
a task instruction, the two camera views, proprioception, and the two tactile views, while vision-only
baselines and the \emph{vision only} ablation drop the tactile views.  The tactile expert runs
asynchronously while a chunk executes, so corrections apply within the same chunk rather than at
the next replanning step.  The supplement reports the action space,
control rate, and model configuration.

\paragraph{Tasks and success criteria.}
The five tasks---charger-plug insertion, multi-object fruit transfer, fragile potato-chip transfer,
empty-bottle uprighting, and expansion-card insertion---exercise tight alignment, regulated grip
force, fragile-object handling, and multi-stage execution.  Uprighting an \emph{empty} bottle is as
much a force-regulation task as a reorientation one: the bottle is light enough that an over-firm
grasp keeps it pinched between the fingers, so it fails to release and settle upright when they
open.  Every episode has a predeclared binary success condition, and damage counts as failure for
fruit and potato-chip manipulation. The supplement gives one
representative image and a complete description per task, including the initial configuration,
reset protocol, and exact success criterion.

\paragraph{Baselines.}
External baselines cover a general VLA ($\pi_{0.5}$)~\citep{intelligence2025pi_}, action-chunking
imitation (ACT)~\citep{zhao2023learning}, a visuo-tactile policy
(VITaL)~\citep{george2025vital}, a world-action model
(LingBot-VA)~\citep{li2026causal}, and two closely related tactile world-action models (Dream-Tac
and T-Rex)~\citep{lou2026dreamtac,niu2026trex}.  All are adapted to the same robot and action
interface and keep their official observation modalities: vision-only methods are not retrofitted
with tactile input, while Dream-Tac and T-Rex retain tactile sensing.

\paragraph{Evaluation protocol.}
For each method on each task we run 20 real-world evaluation trials under the same task
definitions, success criteria, real-world setup, and trial budget, and report the per-task episode
success rate together with its average across tasks.

\subsection{Results and analysis}
\label{sec:exp_results}

\subsubsection{Q1: Overall contact-rich manipulation}
\label{sec:exp_q1}

\paragraph{Overall comparison.}
Figure~\ref{fig:main_results_bar} reports per-task success rates and their average.  \model{}
attains the highest success rate on all five tasks and averages 64\%, 16 points above the strongest
baseline, T-Rex (48\%).  Measured against the best baseline for each task, the margin is largest on
charger-plug insertion (80\% vs.~65\%) and multi-object fruit transfer (65\% vs.~50\%), and
smallest on potato-chip transfer (90\% vs.~85\%), the least demanding task in the suite, where
four baselines already reach 70\% or above.  Difficulty concentrates on the two tightest-tolerance
tasks: expansion-card insertion is never completed by any baseline outside the two tactile
world-action models, and empty-bottle uprighting is never completed by $\pi_{0.5}$ or ACT.
Neither ingredient by itself accounts for this.  VITaL, a reactive visuo-tactile policy (22\%), and
LingBot-VA, a vision-only WAM (28\%), both fall below the vision-only VLA $\pi_{0.5}$ (36\%), and
only Dream-Tac (45\%) and T-Rex (48\%), which combine tactile sensing with world-action modeling,
improve on it. Overall, \model{} combines tactile prediction
with prediction-informed action correction and, even without pretraining, outperforms every prior
method evaluated across all five tasks.

\paragraph{Comparison with tactile world-action models.}
Dream-Tac and T-Rex lead on different groups of tasks, and the split follows what each one
lacks.  Dream-Tac leads on the two tasks whose success criterion counts damage---potato-chip
transfer (85\% vs.~75\%) and fruit transfer (40\% vs.~35\%)---where the grip has to be right
as contact is established, a decision its anticipated contact gives it a prior for.  It falls
behind on the two tight-tolerance insertions, plug (55\% vs.~65\%) and card (20\% vs.~35\%),
in which the misalignment becomes observable only through contact partway into the chunk: a chunk
committed in a single generative pass cannot act on evidence that arrives after it is issued.
T-Rex shows the opposite pattern.  Revising the chunk from the live tactile stream makes it the strongest
baseline on both insertions, but each reading arrives with no statement of what contact the plan
expected, so it responds only once a physical error has been sensed---and on the damage-scored
tasks that is often too late, since a crushed chip is not restored by a later correction.
\model{} leads on both groups---90\% and 65\% on chip and fruit, 80\% and 45\% on
plug and card---as expected if a correction is read against the contact the plan anticipated rather
than against the reading alone.  Q3 removes exactly that access while leaving both abilities in
place.

\subsection{Ablation study}
\label{sec:exp_ablation}

\subsubsection{Q2: Why is tactile prediction alone insufficient?}
\label{sec:exp_q2}

\begin{figure}[t]
\centering
\includegraphics[width=\columnwidth]{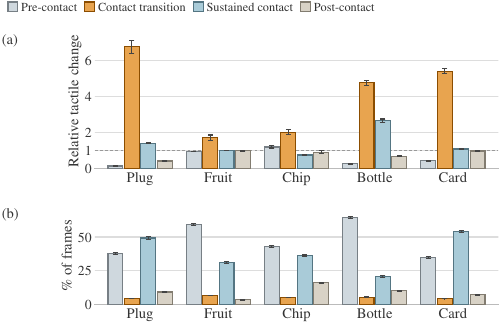}
\caption{\textbf{Tactile analysis by execution phase.}  \textbf{(a)}~Adjacent tactile change
relative to the episode mean (dashed line).  \textbf{(b)}~Proportion of an episode's frames.}
\label{fig:tactile_activity}
\end{figure}

\paragraph{Data-level evidence.}
We measure $c_t$, the mean absolute change between consecutive frames averaged over the two tactile
views, on training episodes of all five tasks, and label execution phases from the measured gripper
width alone (full protocol in the supplement).  Contact transitions are the most active phase in
every task, and also one of the shortest: they hold 4.9\% of the frames but 18.6\% of the total
tactile change (Fig.~\ref{fig:tactile_activity}).  Most of the prediction objective is therefore
dominated by frames in which the tactile image barely changes.  Moreover, the active frames become
available only after execution of the action chunk has begun: they occur within the chunk rather
than before it, by which point the contact-conditioned plan has already been committed.

\paragraph{Evaluation-level evidence.}
Consistent with that limit, adding the prediction objective raises success only part of the
way.  Under matched data, backbone, action interface, optimization steps, and evaluation resets,
predicting tactile observations lifts the base WAM from 22\% to 37\% average success
(Table~\ref{tab:component_ablation}, \emph{vision only} vs.~\emph{w/o tactile expert}), whereas
the full system reaches 64\%: roughly a third of the total gain over the vision-centric WAM is
available from the prediction objective alone.  Prediction can state what contact to expect, but
nothing acts on the discrepancy until a correction path reads it, which Q3 tests.

\subsubsection{Q3: Which components enable tactile action correction?}
\label{sec:exp_q3}

\paragraph{Component ablation design.}
Each variant in Table~\ref{tab:component_ablation} cuts one pathway between tactile sensing and the
executed action.  \emph{Vision only} removes all tactile views and the tactile expert, leaving a
visual--action WAM.  \emph{Without tactile expert} retains the tactile-predictive base model and
its planned chunk but removes real-time correction, isolating tactile prediction alone.  The
remaining two variants keep the tactile expert, and with it the current tactile image, and differ
only in what that image is read against.  \emph{Without tactile prediction} removes tactile views
from both the input and prediction target of the video expert.  The planned chunk is therefore
built from visual observations alone, and no predicted tactile enters the cache.  The retained
expert corrects that visually planned chunk from the current tactile image, making this variant a
reactive tactile corrector on our own backbone.  \emph{Without tactile cache} keeps the prediction
and the plan made under it, but blocks the expert's attention to the cached tactile keys, removing
only its direct access to the anticipated contact.  \model{} retains all paths.

\paragraph{Every pathway is necessary, and the cached prediction is what the corrector needs.}
\model{} reaches 64\% average success, compared with 37\% without the tactile expert and 33\%
without tactile prediction.  Prediction without correction and reactive correction without
prediction perform similarly, and neither approaches the full system, indicating that neither
pathway substitutes for the other.  The cleanest evidence for the cached prediction comes from
\emph{without tactile cache} vs.~ \model{}: the base model, tactile expert, current tactile image,
and parameter count are unchanged.  Only the expert's access to the cached tactile keys differs.
Enabling that access raises success from 47\% to 64\%, supporting our central claim that current
tactile feedback should be read against the anticipated contact.  This interpretation is reinforced
by the reactive variant, which falls below \emph{vision only} on the two damage-sensitive tasks,
chip transfer (45\% vs.~60\%) and fruit transfer (25\% vs.~30\%), suggesting that tactile
feedback without an expectation can pull actions away from an already correct plan.  Finally,
because \emph{without tactile cache} still plans under tactile prediction, its 14-point gain over
\emph{without tactile prediction} measures the value of tactile-predictive planning without direct
cache access.  Granting the corrector that access supplies the remaining 17 points.

\subsubsection{Q4: Can the tactile expert correct a chunk in real time?}
\label{sec:exp_q4}

\begin{table}[t]
\centering
\caption{\textbf{Inference efficiency of \model{}.}  Average wall-clock latency and the
corresponding rate of each inference stage.}
\label{tab:online_efficiency}
\footnotesize
\setlength{\tabcolsep}{3pt}
\begin{tabular}{lcc}
\toprule
Inference stage & Latency (ms) $\downarrow$ & Rate (Hz) $\uparrow$ \\
\midrule
Chunk generation (ten flow steps) & 628.6 & 1.59 \\
Cache prefill (one clean pass) & 62.6 & 15.97 \\
Tactile correction (one pass) & 30.4 & 32.90 \\
\bottomrule
\end{tabular}
\end{table}

As shown in Table~\ref{tab:online_efficiency}, one tactile-expert pass takes only $30.4$~ms,
corresponding to an execution rate of $32.90$~Hz and a $20.7\times$ reduction in latency relative
to regenerating an action chunk.  The tactile expert can therefore respond to incoming tactile
feedback at high frequency.  We further measure execution traces across all five tasks.  In only
$1.6\%$ of tactile-expert calls
does the robot advance by one or more control steps while the correction is being computed.  Almost
all corrections are thus returned before the next step.  Moreover, the asynchronous inference
scheme of Sec.~\ref{sec:method_optimization} keeps the robot executing the latest committed chunk
and writes each correction only to its unexecuted suffix.  We consequently observe no
correction-induced jerks in the executed motion.

\section{Conclusion}
\label{sec:conclusion}

In this paper, we introduce \model{}, a tactile world-action model for contact-rich manipulation.
Unlike prior tactile world-action models, whose predicted contact is fixed once the action chunk is
planned, \model{} turns tactile prediction into real-time correction during execution.  A
tactile expert reads each new tactile image against the layer-wise tactile
cache, correcting the planned action chunk asynchronously.
We evaluate \model{} on five real-world contact-rich manipulation tasks, where it
attains the highest success rate on every task and improves the average by 16 points over the
strongest baseline.  Each correction takes only $30.4$~ms, $20.7\times$ cheaper than replanning.
Overall, our results indicate that tactile prediction and tactile correction are
complementary, and that reading contact feedback against an explicit expectation is a practical
route to closed-loop contact-rich manipulation.

\bibliography{aaai2027}

@article{intelligence2025pi_,
  title   = {{$\pi_{0.5}$}: A Vision-Language-Action Model with Open-World Generalization},
  author  = {{Physical Intelligence} and Black, Kevin and Brown, Noah and Darpinian, James and
             Dhabalia, Karan and Driess, Danny and Esmail, Adnan and Equi, Michael and
             Finn, Chelsea and Fusai, Niccolo and others},
  journal = {arXiv preprint arXiv:2504.16054},
  year    = {2025},
}

@article{yuan2026fast,
  title   = {Fast-WAM: Do World Action Models Need Test-Time Future Imagination?},
  author  = {Yuan, Tianyuan and Dong, Zibin and Liu, Yicheng and Zhao, Hang},
  journal = {arXiv preprint arXiv:2603.16666},
  year    = {2026},
}

@article{li2026causal,
  title   = {Causal World Modeling for Robot Control},
  author  = {Li, Lin and Zhang, Qihang and Luo, Yiming and Yang, Shuai and Wang, Ruilin and
             Han, Fei and Yu, Mingrui and Gao, Zelin and Xue, Nan and Zhu, Xing and others},
  journal = {arXiv preprint arXiv:2601.21998},
  year    = {2026},
}

@inproceedings{george2025vital,
  title     = {VITaL Pretraining: Visuo-Tactile Pretraining for Tactile and Non-Tactile
               Manipulation Policies},
  author    = {George, Abraham and Gano, Selam and Katragadda, Pranav and Farimani, Amir Barati},
  booktitle = {2025 IEEE International Conference on Robotics and Automation (ICRA)},
  pages     = {258--264},
  year      = {2025},
  organization = {IEEE},
}

@article{zhao2023learning,
  title   = {Learning Fine-Grained Bimanual Manipulation with Low-Cost Hardware},
  author  = {Zhao, Tony Z. and Kumar, Vikash and Levine, Sergey and Finn, Chelsea},
  journal = {arXiv preprint arXiv:2304.13705},
  year    = {2023},
}

@article{lambeta2020digit,
  title   = {{DIGIT: A Novel Design for a Low-Cost Compact High-Resolution Tactile Sensor with Application to In-Hand Manipulation}},
  author  = {Mike Lambeta and Po-Wei Chou and Stephen Tian and Brian Yang and Benjamin Maloon and Victoria Rose Most and Dave Stroud and Raymond Santos and Ahmad Byagowi and Gregg Kammerer and Dinesh Jayaraman and Roberto Calandra},
  journal = {arXiv preprint arXiv:2005.14679},
  year    = {2020},
}

@article{donlon2018gelslim,
  title   = {{GelSlim: A High-Resolution, Compact, Robust, and Calibrated Tactile-sensing Finger}},
  author  = {Elliott Donlon and Siyuan Dong and Melody Liu and Jianhua Li and Edward Adelson and Alberto Rodriguez},
  journal = {arXiv preprint arXiv:1803.00628},
  year    = {2018},
}

@article{lepora2021tactileservo,
  title   = {{Pose-Based Tactile Servoing: Controlled Soft Touch using Deep Learning}},
  author  = {Nathan F. Lepora and John Lloyd},
  journal = {arXiv preprint arXiv:2012.02504},
  year    = {2020},
}

@article{hogan2020tactiledexterity,
  title   = {{Tactile Dexterity: Manipulation Primitives with Tactile Feedback}},
  author  = {Francois R. Hogan and Jose Ballester and Siyuan Dong and Alberto Rodriguez},
  journal = {arXiv preprint arXiv:2002.03236},
  year    = {2020},
}

@article{higuera2024sparsh,
  title   = {{Sparsh: Self-supervised touch representations for vision-based tactile sensing}},
  author  = {Carolina Higuera and Akash Sharma and Chaithanya Krishna Bodduluri and Taosha Fan and Patrick Lancaster and Mrinal Kalakrishnan and Michael Kaess and Byron Boots and Mike Lambeta and Tingfan Wu and Mustafa Mukadam},
  journal = {arXiv preprint arXiv:2410.24090},
  year    = {2024},
}

@article{zhao2024t3,
  title   = {{Transferable Tactile Transformers for Representation Learning Across Diverse Sensors and Tasks}},
  author  = {Jialiang Zhao and Yuxiang Ma and Lirui Wang and Edward H. Adelson},
  journal = {arXiv preprint arXiv:2406.13640},
  year    = {2024},
}

@article{yang2024unitouch,
  title   = {{Binding Touch to Everything: Learning Unified Multimodal Tactile Representations}},
  author  = {Fengyu Yang and Chao Feng and Ziyang Chen and Hyoungseob Park and Daniel Wang and Yiming Dou and Ziyao Zeng and Xien Chen and Rit Gangopadhyay and Andrew Owens and Alex Wong},
  journal = {arXiv preprint arXiv:2401.18084},
  year    = {2024},
}

@article{lee2019visiontouch,
  title   = {{Making Sense of Vision and Touch: Self-Supervised Learning of Multimodal Representations for Contact-Rich Tasks}},
  author  = {Michelle A. Lee and Yuke Zhu and Krishnan Srinivasan and Parth Shah and Silvio Savarese and Li Fei-Fei and Animesh Garg and Jeannette Bohg},
  journal = {arXiv preprint arXiv:1810.10191},
  year    = {2018},
}

@article{li2022seehearfeel,
  title   = {{See, Hear, and Feel: Smart Sensory Fusion for Robotic Manipulation}},
  author  = {Hao Li and Yizhi Zhang and Junzhe Zhu and Shaoxiong Wang and Michelle A Lee and Huazhe Xu and Edward Adelson and Li Fei-Fei and Ruohan Gao and Jiajun Wu},
  journal = {arXiv preprint arXiv:2212.03858},
  year    = {2022},
}

@article{huang2024threedvitac,
  title   = {{3D-ViTac: Learning Fine-Grained Manipulation with Visuo-Tactile Sensing}},
  author  = {Binghao Huang and Yixuan Wang and Xinyi Yang and Yiyue Luo and Yunzhu Li},
  journal = {arXiv preprint arXiv:2410.24091},
  year    = {2024},
}

@article{he2024foar,
  title   = {{FoAR: Force-Aware Reactive Policy for Contact-Rich Robotic Manipulation}},
  author  = {Zihao He and Hongjie Fang and Jingjing Chen and Hao-Shu Fang and Cewu Lu},
  journal = {arXiv preprint arXiv:2411.15753},
  year    = {2024},
}

@article{heng2025vitacformer,
  title   = {{ViTacFormer: Learning Cross-Modal Representation for Visuo-Tactile Dexterous Manipulation}},
  author  = {Liang Heng and Haoran Geng and Kaifeng Zhang and Pieter Abbeel and Jitendra Malik},
  journal = {arXiv preprint arXiv:2506.15953},
  year    = {2025},
}

@article{huang2025tactilevla,
  title   = {{Tactile-VLA: Unlocking Vision-Language-Action Model's Physical Knowledge for Tactile Generalization}},
  author  = {Jialei Huang and Shuo Wang and Fanqi Lin and Yihang Hu and Chuan Wen and Yang Gao},
  journal = {arXiv preprint arXiv:2507.09160},
  year    = {2025},
}

@article{lou2026dreamtac,
  title   = {{Dream-Tac: A Unified Tactile World Action Model for Contact-Rich Robot Manipulation}},
  author  = {Yunfan Lou and Yifan Ye and Yankai Fu and Jun Cen and Xiaowei Chi and Yaoxu Lyu and Peidong Jia and Sirui Han and Zhihe Lu and Shanghang Zhang},
  journal = {arXiv preprint arXiv:2606.08737},
  year    = {2026},
}

@article{zang2026tacforesight,
  title   = {{TacForeSight: Force-Guided Tactile World Model for Contact-Rich Manipulation}},
  author  = {Yujie Zang and Yuhang Zheng and Xian Nie and Yupeng Zheng and Shuai Tian and Songen Gu and Chen Gao and Zining Wang and Shuicheng Yan and Wenchao Ding},
  journal = {arXiv preprint arXiv:2606.11184},
  year    = {2026},
}

@article{ye2025dreamtacvla,
  title   = {{Learning to Feel the Future: DreamTacVLA for Contact-Rich Manipulation}},
  author  = {Guo Ye and Zexi Zhang and Xu Zhao and Shang Wu and Haoran Lu and Shihan Lu and Han Liu},
  journal = {arXiv preprint arXiv:2512.23864},
  year    = {2025},
}

@article{zheng2026omnivta,
  title   = {{OmniVTA: Visuo-Tactile World Modeling for Contact-Rich Robotic Manipulation}},
  author  = {Yuhang Zheng and Songen Gu and Weize Li and Yupeng Zheng and Yujie Zang and Shuai Tian and Xiang Li and Ce Hao and Chen Gao and Si Liu and Haoran Li and Yilun Chen and Shuicheng Yan and Wenchao Ding},
  journal = {arXiv preprint arXiv:2603.19201},
  year    = {2026},
}

@article{niu2026touchdreaming,
  title   = {{Learning Versatile Humanoid Manipulation with Touch Dreaming}},
  author  = {Yaru Niu and Zhenlong Fang and Binghong Chen and Shuai Zhou and Revanth Krishna Senthilkumaran and Hao Zhang and Bingqing Chen and Chen Qiu and H. Eric Tseng and Jonathan Francis and Ding Zhao},
  journal = {arXiv preprint arXiv:2604.13015},
  year    = {2026},
}

@article{black2024pi0,
  title   = {{$\pi_0$: A Vision-Language-Action Flow Model for General Robot Control}},
  author  = {Kevin Black and Noah Brown and Danny Driess and Adnan Esmail and Michael Equi and Chelsea Finn and Niccolo Fusai and Lachy Groom and Karol Hausman and Brian Ichter and Szymon Jakubczak and Tim Jones and Liyiming Ke and Sergey Levine and Adrian Li-Bell and Mohith Mothukuri and Suraj Nair and Karl Pertsch and Lucy Xiaoyang Shi and James Tanner and Quan Vuong and Anna Walling and Haohuan Wang and Ury Zhilinsky},
  journal = {arXiv preprint arXiv:2410.24164},
  year    = {2024},
}

@article{tang2023industreal,
  title   = {{IndustReal: Transferring Contact-Rich Assembly Tasks from Simulation to Reality}},
  author  = {Bingjie Tang and Michael A. Lin and Iretiayo Akinola and Ankur Handa and Gaurav S. Sukhatme and Fabio Ramos and Dieter Fox and Yashraj Narang},
  journal = {arXiv preprint arXiv:2305.17110},
  year    = {2023},
}

@article{zhang2023admittance,
  title   = {{Efficient Sim-to-real Transfer of Contact-Rich Manipulation Skills with Online Admittance Residual Learning}},
  author  = {Xiang Zhang and Changhao Wang and Lingfeng Sun and Zheng Wu and Xinghao Zhu and Masayoshi Tomizuka},
  journal = {arXiv preprint arXiv:2310.10509},
  year    = {2023},
}

@article{niu2026trex,
  title   = {{T-Rex: Tactile-Reactive Dexterous Manipulation}},
  author  = {Dantong Niu and Zhuoyang Liu and Zekai Wang and Boning Shao and Zhao-Heng Yin and Anirudh Pai and Yuvan Sharma and Stefano Saravalle and Ruijie Zheng and Jing Wang and Ryan Punamiya and Mengda Xu and Yuqi Xie and Yunfan Jiang and Letian Fu and Konstantinos Kallidromitis and Matteo Gioia and Junyi Zhang and Jiaxin Ge and Haiwen Feng and Fabio Galasso and Wei Zhan and David M. Chan and Yutong Bai and Roei Herzig and Jiahui Lei and Li Fei-Fei and Ken Goldberg and Jitendra Malik and Pieter Abbeel and Yuke Zhu and Danfei Xu and Linxi Fan and Trevor Darrell},
  journal = {arXiv preprint arXiv:2606.17055},
  year    = {2026},
}

@article{yu2025forcevla,
  title   = {{ForceVLA: Enhancing VLA Models with a Force-aware MoE for Contact-rich Manipulation}},
  author  = {Jiawen Yu and Hairuo Liu and Qiaojun Yu and Jieji Ren and Ce Hao and Haitong Ding and Guangyu Huang and Guofan Huang and Yan Song and Panpan Cai and Cewu Lu and Wenqiang Zhang},
  journal = {arXiv preprint arXiv:2505.22159},
  year    = {2025},
}

@article{kim2024openvla,
  title   = {{OpenVLA: An Open-Source Vision-Language-Action Model}},
  author  = {Moo Jin Kim and Karl Pertsch and Siddharth Karamcheti and Ted Xiao and Ashwin Balakrishna and Suraj Nair and Rafael Rafailov and Ethan Foster and Grace Lam and Pannag Sanketi and Quan Vuong and Thomas Kollar and Benjamin Burchfiel and Russ Tedrake and Dorsa Sadigh and Sergey Levine and Percy Liang and Chelsea Finn},
  journal = {arXiv preprint arXiv:2406.09246},
  year    = {2024},
}

@article{zhang2025vtla,
  title   = {{VTLA: Vision-Tactile-Language-Action Model with Preference Learning for Insertion Manipulation}},
  author  = {Chaofan Zhang and Peng Hao and Xiaoge Cao and Xiaoshuai Hao and Shaowei Cui and Shuo Wang},
  journal = {arXiv preprint arXiv:2505.09577},
  year    = {2025},
}

@article{brohan2023rt2,
  title   = {{RT-2: Vision-Language-Action Models Transfer Web Knowledge to Robotic Control}},
  author  = {Anthony Brohan and Noah Brown and Justice Carbajal and Yevgen Chebotar and Xi Chen and Krzysztof Choromanski and Tianli Ding and Danny Driess and Avinava Dubey and Chelsea Finn and Pete Florence and Chuyuan Fu and Montse Gonzalez Arenas and Keerthana Gopalakrishnan and Kehang Han and Karol Hausman and Alexander Herzog and Jasmine Hsu and Brian Ichter and Alex Irpan and Nikhil Joshi and Ryan Julian and Dmitry Kalashnikov and Yuheng Kuang and Isabel Leal and Lisa Lee and Tsang-Wei Edward Lee and Sergey Levine and Yao Lu and Henryk Michalewski and Igor Mordatch and Karl Pertsch and Kanishka Rao and Krista Reymann and Michael Ryoo and Grecia Salazar and Pannag Sanketi and Pierre Sermanet and Jaspiar Singh and Anikait Singh and Radu Soricut and Huong Tran and Vincent Vanhoucke and Quan Vuong and Ayzaan Wahid and Stefan Welker and Paul Wohlhart and Jialin Wu and Fei Xia and Ted Xiao and Peng Xu and Sichun Xu and Tianhe Yu and Brianna Zitkovich},
  journal = {arXiv preprint arXiv:2307.15818},
  year    = {2023},
}

@article{liang2025mixtureoftransformers,
  title   = {Mixture-of-Transformers: A Sparse and Scalable Architecture for Multi-Modal Foundation Models},
  author  = {Liang, Weixin and Yu, Lili and Luo, Liang and Iyer, Srini and Dong, Ning and
             Zhou, Chunting and Ghosh, Gargi and Lewis, Mike and Yih, Wen-tau and
             Zettlemoyer, Luke and Lin, Xi Victoria},
  journal = {Transactions on Machine Learning Research},
  year    = {2025},
}

@article{yu2026wall,
  title   = {{Wall-oss-0.5 technical report}},
  author  = {Ryan Yu and Pushi Zhang and Starrick Liu and Brae Liu and Miracle Kang and Shalfun Li and Lights Shi and Ellie Ma and Ping Yang and Chris Pan and others},
  journal = {arXiv preprint arXiv:2605.30877},
  year    = {2026},
}

@article{qwenvla,
  title   = {{Qwen-VLA: Unifying Vision-Language-Action Modeling across Tasks, Environments, and Robot Embodiments}},
  author  = {Qiuyue Wang and Mingsheng Li and Jian Guan and Jinhui Ye and Sicheng Xie and Yitao Liu and Junhao Chen and Zhixuan Liang and Jie Zhang and Xintong Hu and Xuhong Huang and Pei Lin and Junyang Lin and Dayiheng Liu and Shuai Bai and Jingren Zhou and Jiazhao Zhang and Haoqi Yuan and Gengze Zhou and Hang Yin and Ye Wang and Yiyang Huang and Zixing Lei and Wujian Peng and Delin Chen and Yingming Zheng and Jingyang Fan and Xianwei Zhuang and Xin Zhou and Haoyang Li and Anzhe Chen and Tong Zhang and Xuejing Liu and Yuchong Sun and Ruizhe Chen and Zhaohai Li and Chenxu Lü and Zhibo Yang and Tao Yu and Xionghui Chen},
  journal = {arXiv preprint arXiv:2605.30280},
  year    = {2026},
}

@article{yuan2026qwen,
  title   = {{Qwen-robotmanip technical report: Alignment unlocks scale for robotic manipulation foundation models}},
  author  = {Haoqi Yuan and Zhixuan Liang and Anzhe Chen and Ye Wang and Haoyang Li and Pei Lin and Yiyang Huang and Zixing Lei and Tong Zhang and Jiazhao Zhang and others},
  journal = {arXiv preprint arXiv:2606.17846},
  year    = {2026},
}

@article{wu2026foundation,
  title   = {{From Foundation to Application: Improving VLA Models in Practice}},
  author  = {Wei Wu and Fangjing Wang and Fan Lu and He Sun and Shi Liu and Yunnan Wang and Yibin Yan and Yong Wang and Shuailei Ma and Xinyang Wang and others},
  journal = {arXiv preprint arXiv:2607.06403},
  year    = {2026},
}

@article{zhang2026hy,
  title   = {{Hy-Embodied-0.5-VLA: From Vision-Language-Action Models to a Real-World Robot Learning Stack}},
  author  = {He Zhang and Lingzhu Xiang and Haitao Lin and Zeyu Huang and Minghui Wang and Dingyan Zhong and Yubo Dong and Yihao Wu and Yongming Rao and Dongsheng Zhang and others},
  journal = {arXiv preprint arXiv:2606.14409},
  year    = {2026},
}

@article{li2026wall,
  title   = {{WALL-WM: Carving World Action Modeling at the Event Joints}},
  author  = {Shalfun Li and Victor Yao and Charles Yang and Truth Qu and Regis Cheng and Ryan Yu and Howard Lu and Newton Von and Vincent Chen and Yohann Tang and others},
  journal = {arXiv preprint arXiv:2606.01955},
  year    = {2026},
}

@article{ye2026world,
  title   = {{World action models are zero-shot policies}},
  author  = {Seonghyeon Ye and Yunhao Ge and Kaiyuan Zheng and Shenyuan Gao and Sihyun Yu and George Kurian and Suneel Indupuru and You Liang Tan and Chuning Zhu and Jiannan Xiang and others},
  journal = {arXiv preprint arXiv:2602.15922},
  year    = {2026},
}

@article{kim2026cosmos,
  title   = {{Cosmos policy: Fine-tuning video models for visuomotor control and planning}},
  author  = {Moo Jin Kim and Yihuai Gao and Tsung-Yi Lin and Yen-Chen Lin and Yunhao Ge and Grace Lam and Percy Liang and Shuran Song and Ming-Yu Liu and Chelsea Finn and others},
  journal = {arXiv preprint arXiv:2601.16163},
  year    = {2026},
}

@article{ma2026dit4dit,
  title   = {{DiT4DiT: Jointly Modeling Video Dynamics and Actions for Generalizable Robot Control}},
  author  = {Teli Ma and Jia Zheng and Zifan Wang and Chunli Jiang and Andy Cui and Junwei Liang and Shuo Yang},
  journal = {arXiv preprint arXiv:2603.10448},
  year    = {2026},
}

@article{zhang2026imagewam,
  title   = {{ImageWAM: Do World Action Models Really Need Video Generation, or Just Image Editing?}},
  author  = {Yuyang Zhang and Wenyao Zhang and Zekun Qi and He Zhang and Haitao Lin and Jingbo Zhang and Yao Mu and Xiaokang Yang and Wenjun Zeng and Xin Jin},
  journal = {arXiv preprint arXiv:2606.19531},
  year    = {2026},
}

@article{yuan2026vtam,
  title={Vtam: Video-tactile-action models for complex physical interaction beyond vlas},
  author={Yuan, Haoran and Yi, Weigang and Zhang, Zhenyu and Chen, Wendi and Mo, Yuchen and Yin, Jiashi and Li, Xinzhuo and Zeng, Xiangyu and Wen, Chuan and Lu, Cewu and others},
  journal={arXiv preprint arXiv:2603.23481},
  year={2026}
}

@article{n0foundation2026,
  title   = {{N0-Foundation}: Towards the Age of Tactile Intelligence},
  author  = {{NeoteAI Team} and {Fudan TEAI Team}},
  journal = {Technical Report},
  year    = {2026},
}

@article{n0twam2026,
  title   = {{N0-TWAM}: Scaling Tactile-Native World Action Model for Contact-Rich Manipulation},
  author  = {{NeoteAI Team} and {Fudan TEAI Team}},
  journal = {Technical Report},
  year    = {2026},
}

@article{team2026xiaomi,
  title={Xiaomi-Robotics-1: Scaling Vision-Language-Action Models with over 100K Hours of Real-World Trajectories},
  author={Team, Xiaomi Robotics and Guo, Jun and Jin, Piaopiao and Li, Jason and Li, Peiyan and Li, Yingyan and Liu, Futeng and Peng, Wanli and Qin, Optimus and Su, Yifei and others},
  journal={arXiv preprint arXiv:2607.15330},
  year={2026}
}

@article{yuan2026ftp,
  title={Ftp-1: A generalist foundation tactile policy across tactile sensors for contact-rich manipulation},
  author={Yuan, Chengbo and Zhang, Zicheng and Zhou, Mingjie and Chen, Wendi and Wang, Yi and Liu, Zhuoyang and Niu, Dantong and Wang, Shuo and Zhang, Hui and Zhang, Wenkang and others},
  journal={arXiv preprint arXiv:2606.13102},
  year={2026}
}

@article{xue2025reactive,
  title={Reactive diffusion policy: Slow-fast visual-tactile policy learning for contact-rich manipulation},
  author={Xue, Han and Ren, Jieji and Chen, Wendi and Zhang, Gu and Fang, Yuan and Gu, Guoying and Xu, Huazhe and Lu, Cewu},
  journal={arXiv preprint arXiv:2503.02881},
  year={2025}
}

\clearpage
\appendix
\section*{Appendix}
\section{Notation}
\label{sec:supp_notation}

Table~\ref{tab:notation} summarizes the notation used in the main paper.  Throughout, flow time
$\tau$ always denotes the noise level of the conditional flow, whereas physical time $t$
indexes execution stages along the action horizon.  An undecorated symbol denotes a clean
(ground-truth) quantity, a subscript $\tau$ marks its noisy sample at flow time $\tau$, and a
hat marks a model estimate.  Bold lower-case symbols denote vectors or vector-valued signals, and
italic letters denote scalars and indices.

\begin{table*}[t]
\centering
\small
\begin{tabular}{@{}lll@{}}
\toprule
Symbol & Meaning & Current configuration \\
\midrule
\multicolumn{3}{@{}l}{\emph{Observations, actions, and flow matching}} \\
$\mathbf{I}_t$ & Raw tactile image at time $t$ & -- \\
$\mathbf{a}$, $H$ & Demonstrated (clean) action chunk and its horizon & $H=48$ \\
$\hat{\mathbf{a}}$ & Planned action chunk (base-model output, before correction) & -- \\
$\mathbf{z}$ & Clean video latent covering visual and tactile views & -- \\
$\mathbf{x}$, $\boldsymbol{\epsilon}$, $\mathbf{x}_{\tau}$ & Clean sample, Gaussian noise, and noisy interpolant & -- \\
$\tau$ & Flow time (noise level) & $\tau\in[0,1]$ \\
$\mathbf{u}$ & Target flow velocity $\boldsymbol{\epsilon}-\mathbf{x}$ & -- \\
$\mathbf{z}_{\tau}$, $\mathbf{a}_{\tau}$ & Noisy video latent and action chunk at flow time $\tau$ & -- \\
\midrule
\multicolumn{3}{@{}l}{\emph{Tactile correction}} \\
$t$ & Physical-time (execution-stage) index & -- \\
$i$ & Action position within the chunk & $i\in\{0,\ldots,H-1\}$ \\
$m$ & Number of already executed actions (correction offset) & $m\in[0,H)$ \\
$\mathcal{C}$ & Per-layer tactile and action key/value cache built by the prefill & -- \\
$f$ & Tactile expert & -- \\
$\Delta\mathbf{a}$ & Delta action, yielding corrected chunk $\hat{\mathbf{a}}+\Delta\mathbf{a}$ & -- \\
$d$ & Delay margin before which a correction is not written back & action steps \\
\midrule
\multicolumn{3}{@{}l}{\emph{Training}} \\
$\mathcal{L}_{\mathrm{tac}}$ & Stage-2 tactile correction loss & -- \\
$K$ & Correction offsets sampled per planned chunk & $K=4$ \\
\midrule
\multicolumn{3}{@{}l}{\emph{Experimental diagnostics}} \\
$c_t$ & Mean absolute tactile-image change between consecutive frames & -- \\
$n$ & Tactile view index & $n\in\{1,2\}$ \\
\bottomrule
\end{tabular}
\caption{\textbf{Notation.}  Symbols used in the main paper and in this supplement, grouped by
role.  The last column lists the value in the current configuration where one is fixed, while ``--''
marks symbols whose value is context-dependent.}
\label{tab:notation}
\end{table*}

\section{Implementation details}
\label{sec:supp_implementation}

\paragraph{Model configuration.}
The video, action, and tactile experts share 30 Transformer layers and identical attention
geometry (24 heads of dimension 128), so the tactile expert can consume the base experts'
per-layer caches.  Their hidden and feed-forward widths are 3072/14336, 1024/4096, and 1024/4096,
respectively.  A linear bridge maps the video-frontend width (3072) to the tactile-expert width
(1024).  The tactile expert is initialized from the trained action expert, whose blocks it matches
except for the absent cross-attention.  Its output head is zero-initialized so that the first
correction is the identity.

\paragraph{Flow schedule and temporal layout.}
Chunk generation uses ten flow steps, after which one additional clean pass at $\tau=0$ builds the
cache.  The action horizon is $H=48$, and each action comprises a seven-dimensional joint command,
expressed relative to the joint configuration at the first step of the chunk, and a
one-dimensional gripper command.  Commands are issued at 30~Hz, so one chunk spans $1.6$~s of
robot motion and is executed to its full horizon before the next chunk is planned.  The video tokenizer has a VAE temporal stride of 4 and a
future-frame stride of 4, so one latent frame covers 16 consecutive actions and the horizon spans
three video latent frames.  Visual views are rendered at $256\times256$ and tactile views at
$128\times128$ on the shared canvas.

\paragraph{Stage-2 training.}
Each frozen plan and prefill is shared by $K=4$ correction offsets, whose losses are
averaged.  Stage~2 freezes the video expert, action expert, VAE, text encoder, and proprioceptive
encoder.  Only the tactile expert is trainable.  It is optimized at a base learning rate of
$10^{-4}$ under a cosine schedule.

\paragraph{Optimization and computing infrastructure.}
Both stages are trained on a single node with eight NVIDIA H100 GPUs, using a global batch size of
64 for 10 epochs.  The released implementation requires Python~3.10 or newer and pins
\texttt{transformers}~4.57.0, \texttt{accelerate}~1.12.0, and \texttt{DeepSpeed}~0.16.9.

\begin{figure}[t]
\centering
\includegraphics[width=\columnwidth]{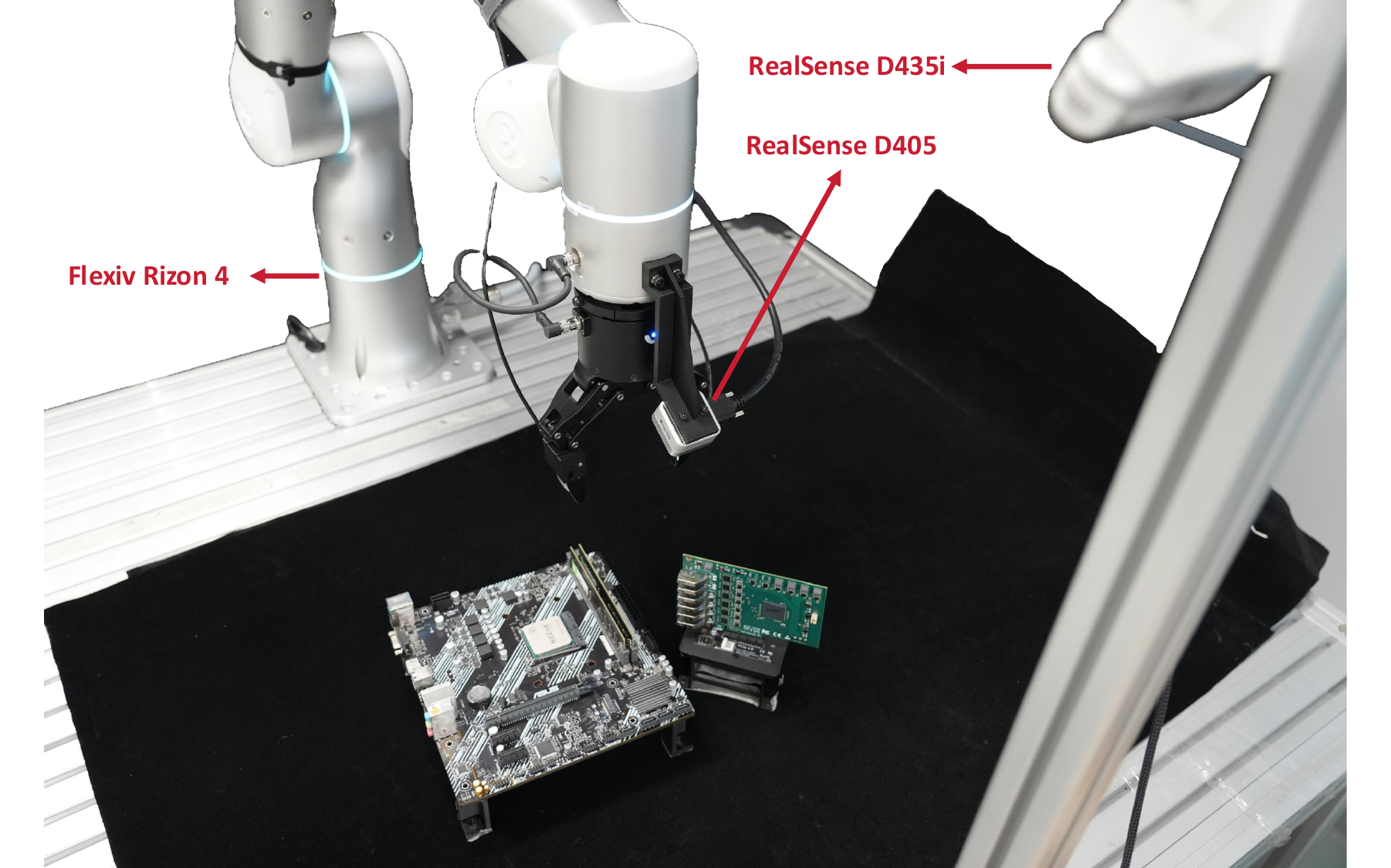}
\caption{\textbf{Real-world experimental setup.}  We use a Flexiv Rizon~4 robot equipped with a
wrist-mounted Intel RealSense D405 camera and a third-person Intel RealSense D435i camera.  The
gripper integrates two InTac~S1 tactile sensors.}
\label{fig:exp_setup}
\end{figure}

\FloatBarrier

\section{Evaluation details}
\label{sec:supp_evaluation}

\paragraph{Trial accounting and task reset.}
For every method--task pair, all 20 attempted evaluation trials are included in the reported
success rate.  Any failure to satisfy the task's binary success criterion---including unsuccessful
grasping, incomplete insertion or transfer, object damage where prohibited, or failure to complete
the task---is counted as a failure; no trial is excluded or repeated based on its outcome.  Within
each task, every method is evaluated with the same reset procedure, task-specific initial
configuration, trial budget, and binary success criterion.  Before every trial, the robot and task
objects are returned to the task's stated starting configuration without any method-specific
adjustment.

\paragraph{Baseline adaptation.}
We train all baselines using the hyperparameters recommended by their official implementations.
For Dream-Tac, we additionally adjust its gating hyperparameters to adapt it to our tactile sensors.

\section{Task descriptions}
\label{sec:supp_tasks}

\begin{figure*}[!t]
\centering
\includegraphics[width=\textwidth]{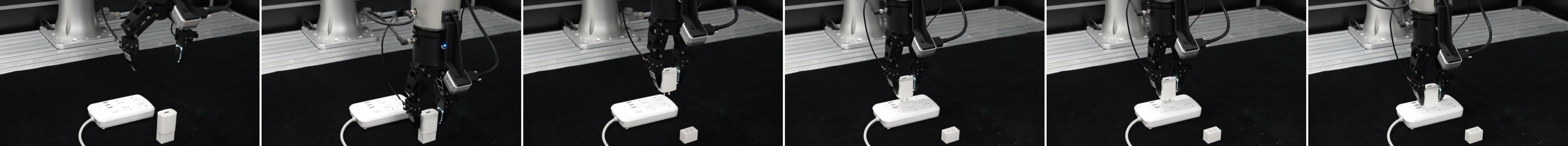}

\vspace{0.1em}
\includegraphics[width=\textwidth]{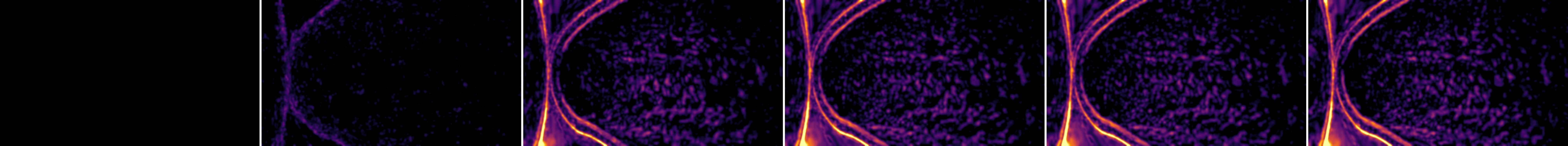}

\vspace{0.15em}
\input{images/trace_legend}

\vspace{0.15em}
\includegraphics[width=0.50\textwidth]{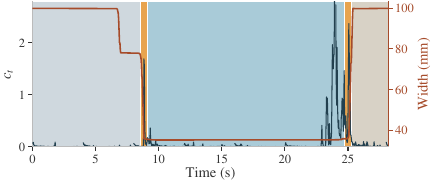}
\caption{\textbf{Charger plug insertion.}  At reset, an unplugged charger adapter stands next to
the power strip.  The robot must grasp the adapter, align its prongs with the designated outlet,
and press it into the socket; an episode succeeds when the plug is seated in the target outlet.}
\label{fig:trace_plug}
\end{figure*}

Each task is described below and paired with the tactile-activity trace of one training episode,
so that the phase structure aggregated by the Q2 diagnostic can be inspected task by task.  The
displayed episode is the one whose per-phase activity profile is closest to that task's median,
selected before inspecting any trace.  In every figure, the upper two rows show six time-aligned
visual and tactile observations from a separate successful rollout, with each tactile frame placed
directly below its visual counterpart.  The bottom plot shows tactile activity $c_t$ (dark) and
measured gripper width (red) from the median-profile episode, so it is not frame-aligned with the
rollout above.  Background colors denote the execution phases defined in Sec.~\ref{sec:supp_q2};
the vertical scale of $c_t$ differs between tasks because the stored videos differ in bitrate.

\begin{figure*}[!t]
\centering
\includegraphics[width=\textwidth]{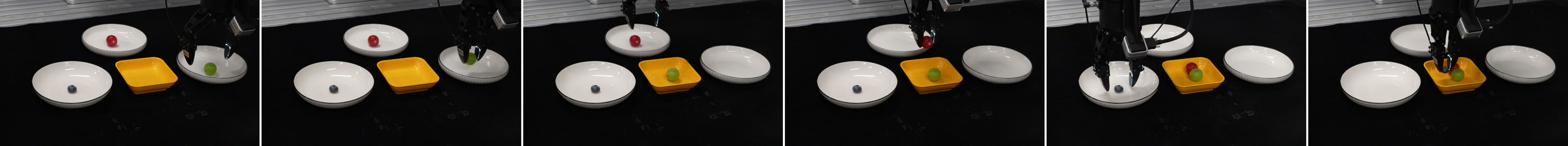}

\vspace{0.1em}
\includegraphics[width=\textwidth]{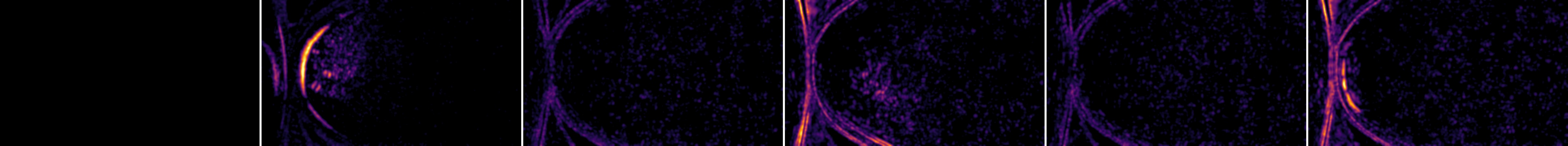}

\vspace{0.15em}
\input{images/trace_legend}

\vspace{0.15em}
\includegraphics[width=0.50\textwidth]{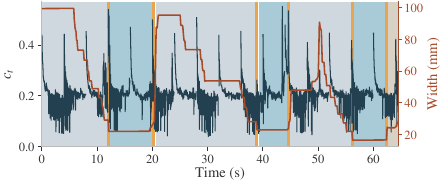}
\caption{\textbf{Multi-object fruit transfer.}  At reset, one fruit is placed on each of three
source plates surrounding an empty yellow tray.  The robot must transfer the fruits sequentially
into the tray; an episode succeeds only when all three fruits are in the tray without visible
damage.}
\label{fig:trace_fruit}
\end{figure*}

\begin{figure*}[!t]
\centering
\includegraphics[width=\textwidth]{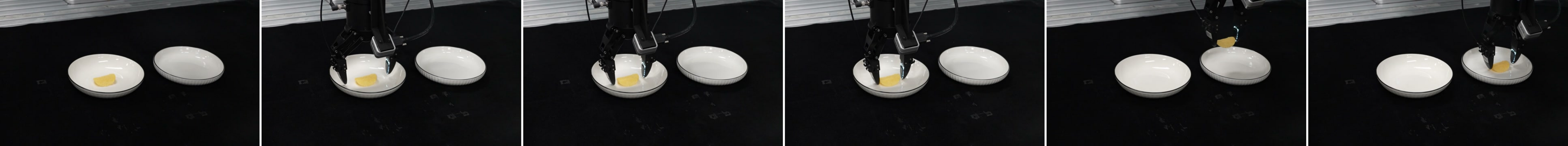}

\vspace{0.1em}
\includegraphics[width=\textwidth]{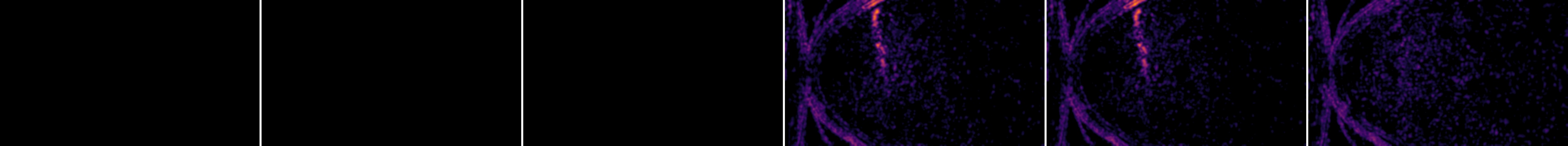}

\vspace{0.15em}
\input{images/trace_legend}

\vspace{0.15em}
\includegraphics[width=0.50\textwidth]{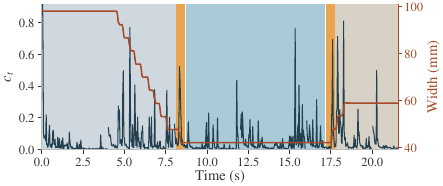}
\caption{\textbf{Fragile potato-chip pick-and-place.}  At reset, an intact potato chip lies in the
source plate and the target plate is empty.  The robot must grasp the chip without crushing it,
carry it between the plates, and release it into the target; an episode succeeds only when the chip
reaches the target plate without visible damage.}
\label{fig:trace_chip}
\end{figure*}

\begin{figure*}[!t]
\centering
\includegraphics[width=\textwidth]{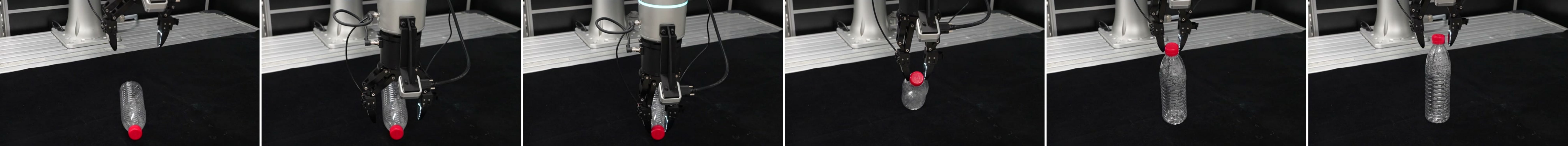}

\vspace{0.1em}
\includegraphics[width=\textwidth]{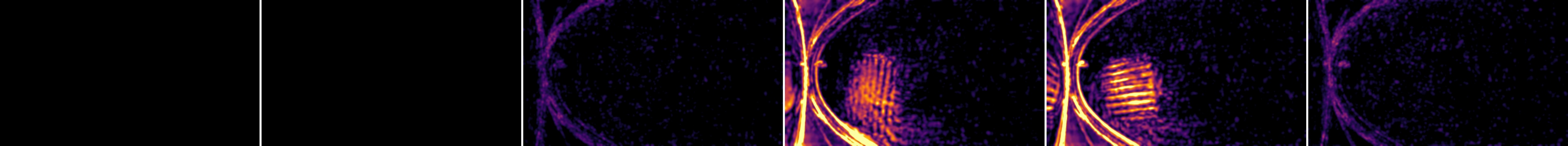}

\vspace{0.15em}
\input{images/trace_legend}

\vspace{0.15em}
\includegraphics[width=0.50\textwidth]{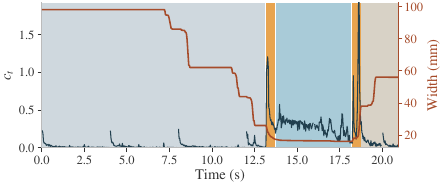}
\caption{\textbf{Empty bottle uprighting.}  At reset, a capped empty plastic bottle lies on its
side.  The robot must grasp the compliant bottle, rotate it to a vertical pose, place its base on
the table, and release it; an episode succeeds only if the bottle remains stably upright after
release.}
\label{fig:trace_bottle}
\end{figure*}

\begin{figure*}[!t]
\centering
\includegraphics[width=\textwidth]{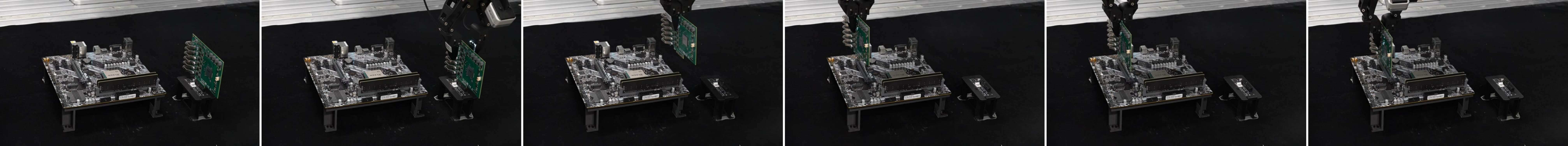}

\vspace{0.1em}
\includegraphics[width=\textwidth]{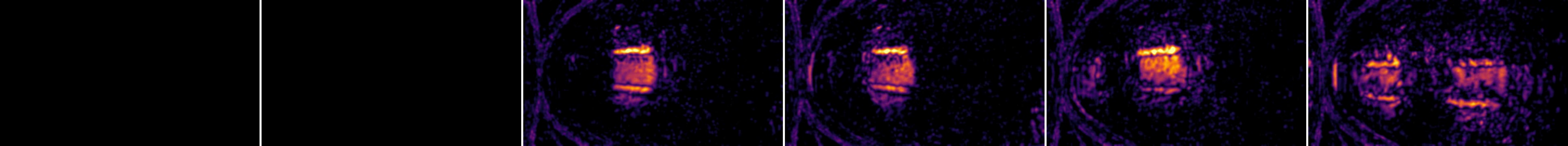}

\vspace{0.15em}
\input{images/trace_legend}

\vspace{0.15em}
\includegraphics[width=0.50\textwidth]{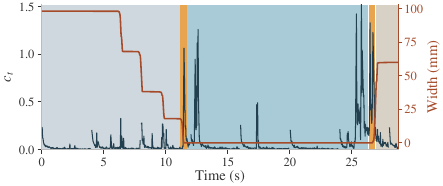}
\caption{\textbf{Expansion-card insertion.}  At reset, the motherboard is fixed in the workspace
and the expansion card stands beside the target slot.  The robot must grasp the card, align its
edge connector with the slot, and press it into the interface; an episode succeeds when the card
is seated in the target slot.}
\label{fig:trace_card}
\end{figure*}

\FloatBarrier
\clearpage

\raggedbottom

\section{Diagnostic for Q2}
\label{sec:supp_q2}

This section gives the protocol behind the diagnostic reported in Q2 of the main paper.

\paragraph{Tactile temporal activity.}
The diagnostic quantifies when tactile observations change, instead of assuming that their useful
content is sparse.  For the two tactile views we measure the per-frame mean absolute change

\begin{equation}
c_t = \frac{1}{2}
\sum_{n=1}^{2}
\operatorname{mean}\left(\left|\mathbf{I}_{t,n}
-\mathbf{I}_{t-1,n}\right|\right),
\label{eq:tactile_change}
\end{equation}

where $n$ indexes the two tactile views and the mean is taken over all pixels.  The diagnostic
covers 296 episodes, about 60 per task, drawn at a fixed stride from each task's episode list.
After the exclusions below, 284{,}948 frames remain.

\paragraph{Full result.}
Activity inside the contact-transition windows is $4.1\pm0.1$ times the episode mean, against
$0.58\pm0.03$ while the fingers are still travelling through free space, $1.36\pm0.04$ during
sustained contact and $0.77\pm0.03$ after release.  The ordering is not an average over a few
extreme episodes: contact-transition activity exceeds the sustained-contact level in 99\% of
episodes and the approach level in 94\%.  The effect is weakest on the multi-object fruit task
($1.7\times$ the episode mean) and on chip transfer ($2.0\times$), and strongest on the two
tight-tolerance insertions and bottle uprighting ($4.9$ to $6.6\times$).

\paragraph{Codec artifacts.}
The tactile streams are stored as HEVC video with a keyframe every 120 frames.  The difference
into a keyframe and the difference out of it are reconstruction discontinuities rather than sensor
change, and are two orders of magnitude larger than a typical frame difference, so both are
dropped, removing 1.7\% of frames.  Bitrate varies across episodes, from 25 to 152~kbit/s, so
$c_t$ is not comparable between episodes in absolute terms.  Every reported contrast is therefore
formed within one episode, by dividing a phase mean by that episode's own mean $c_t$, and episodes
are the unit over which means and standard errors are taken.  Because a low bitrate quantises away
small changes, absolute activity is understated, and understated most in the quietest phases.
Repeating the measurement over a five-frame lag, which lets real deformation accumulate well
beyond the quantiser step, leaves the result unchanged: contact-transition windows then hold
5.0\% of frames and 18.5\% of the total change, with activity $4.00\pm0.18$ times the episode
mean.

\paragraph{Execution-phase annotation.}
Phases are derived from the measured gripper width $w_t$ alone, a signal independent of the
tactile stream.  Teleoperated closures are staircases: the operator shuts the fingers in steps
separated by pauses of up to a second, so neither a width threshold nor "narrowed and stationary"
separates a pause in free space from an arrest against an object.  What separates them is what
follows.  We median-filter $w_t$, mark frames whose finger speed exceeds 10~mm/s as movements, and
label each movement as closing or opening by its net travel, discarding movements shorter than
3~mm.  A \emph{hold} runs from the end of a closing movement to the start of the next opening
movement, and holds shorter than 15 frames are discarded.  A pause partway through a staircase is
followed by more closing rather than by a release, and so is not a hold.  This uses no width
threshold, which is what lets it segment the multi-object task, where the gripper reopens only
partway between objects.  Contact-transition windows are the 8 frames on either side of each hold
boundary, sustained contact is the remainder of a hold, approach is everything before the first
hold, and post-contact everything after the last one.  Contact made earlier in a closure, before
the fingers stop travelling, is labelled approach, and contact between a held object and the
environment is not marked at all.  Both choices work against finding a contact effect rather than
for it.

\end{document}